\documentclass[letterpaper, 10pt, conference]{ieeeconf}

\let\labelindent\relax

\usepackage[inline]{enumitem}
\usepackage{amsmath}
\usepackage{makecell}
\usepackage{multirow}
\usepackage{graphicx}
\usepackage{hyperref}

\usepackage{caption}
\IEEEoverridecommandlockouts

\begin{document}

\title{\LARGE \bf Evaluation of Runtime Monitoring for UAV Emergency Landing}

\author{Joris Guerin$^{1, 2, 3}$, Kevin Delmas$^{3}$ and Jérémie Guiochet$^{1, 2}$
\thanks{$^{1}$Université de Toulouse, $^{2}$LAAS-CNRS, $^{3}$ONERA, Toulouse, France. {\tt\small joris.guerin@laas.fr, kevin.delmas@onera.fr, jeremie.guiochet@laas.fr}}}

\maketitle
\thispagestyle{empty}
\pagestyle{empty}

\begin{abstract}

To certify UAV operations in populated areas, risk mitigation strategies -- such as Emergency Landing (EL) -- must be in place to account for potential failures. EL aims at reducing ground risk by finding safe landing areas using on-board sensors. 
The first contribution of this paper is to present a new EL approach, in line with safety requirements introduced in recent research. In particular, the proposed EL pipeline includes mechanisms to monitor learning based components during execution. This way, another contribution is to study the behavior of Machine Learning Runtime Monitoring (MLRM) approaches within the context of a real-world critical system. A new evaluation methodology is introduced, and applied to assess the practical safety benefits of three MLRM mechanisms. The proposed approach is compared to a default mitigation strategy (open a parachute when a failure is detected), and appears to be much safer.

\end{abstract}

\section{INTRODUCTION}
Unmanned Aerial Vehicles (UAV) can be used in a variety of domains such as construction or structures visual inspection~\cite{drones_urban_environment}. 
The Joint Authorities for Rulemaking of Unmanned Systems recently released a document called Specific Operations Risk Assessment (SORA)~\cite{SORA}, providing guidelines for the development and certification of safe UAV operations. However, applying the detailed risk evaluation and mitigation procedure of the SORA does not allow to certify many useful UAV operations in urban environments~\cite{ssiv}. This is mostly due to the risk of ground impact with third party in case of failure, which imposes very high integrity levels on urban UAVs. A potential solution 
is to introduce new ground risk mitigation strategies, such as Emergency Landing (EL). EL consists in using either external databases or on-board sensors to identify a safe landing area, and to reach it in a potentially degraded control mode. This paper deals with the former sub-problem, sometimes called EL spot detection~\cite{ELspot}, which is simply referred to as EL from now on. 

A review of recent EL approaches is presented in Section~\ref{sec:literature_uav}, and shows that 
there is still no consensus about EL objectives. 
Recent research proposed to clearly define EL objectives for urban UAV, in the form of SORA requirements~\cite{ssiv}. These requirements define the kinds of authorized areas for landing as well as the protocol for testing EL approaches. It also imposes the use of safety monitoring techniques for complex Machine Learning (ML) based components. The first contribution of this paper is to introduce a new approach to urban EL, presented in Section~\ref{sec:approach}, aiming to comply with such requirements. In short, we use a Convolutional Neural Network (CNN) to conduct Semantic Segmentation (SemSeg) on images collected from UAV on-board cameras. The segmentation maps are then processed to find valid landing zones. 
Finally, three runtime monitoring mechanisms from the literature are implemented to gain confidence in the final decisions.

Another contribution is to use EL as a case study to evaluate Machine Learning Runtime Monitoring (MLRM) techniques within the practical context of a critical autonomous system. A literature review about existing MLRM mechanisms (Section~\ref{sec:literature_rm}) shows that their performance is rarely studied in terms of added safety at system level. Hence, a generic evaluation methodology is proposed (Section~\ref{sec:eval_rm}) to assess the safety benefits of MLRM approaches with respect to the whole system. 
This methodology is used to assess the quality of the proposed EL system, and especially of the three different MLRM approaches. 
The results obtained (Section~\ref{sec:results}) represent a significant improvement compared to the default SORA mitigation strategy, i.e., using a parachute to decrease the UAV kinetic energy. 

\begin{figure*}
    \centering
    \includegraphics[width=\textwidth]{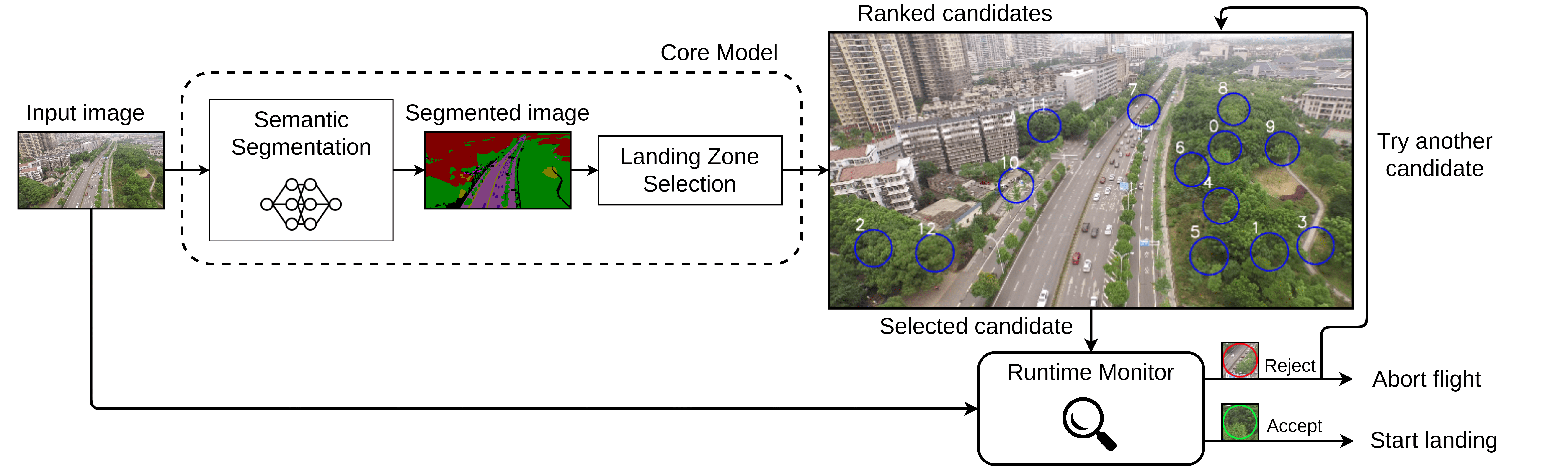}
    \caption{\textbf{Proposed architecture for Emergency Landing}. The Core Model (CM) uses Semantic Segmentation to identify safe landing areas, while the Runtime Monitor (RM) is in charge of detecting CM errors to enhance the safety of the system.}
    \label{fig:overview}
\end{figure*}

\section{RELATED WORK} \label{sec:literature}
\subsection{UAV Emergency Landing} \label{sec:literature_uav}
This literature review focuses on EL techniques using images from on-board cameras. They can be divided into two main lines of work: \begin{enumerate*} \item low-altitude approaches try to identify the right time to trigger a short range landing manoeuvre~\cite{low_altitude_flat, landmark_and_obstacle}, and \item high-altitude approaches aim at finding a distant landing spot that will remain safe for a longer time range \end{enumerate*}. This paper fits into the latter category, which is presented in further details. A first approach identifies areas in the image with low concentration of edges, and classifies them into building, bitumen, trees, grass or water using a Support Vector Machine algorithm~\cite{edges_svm, edges_from_high_altitude_2, edges_from_high_altitude_1}.
Similarly, another technique estimates depth in RGB images using a CNN, and selects flat surfaces for landing (roof-tops, roads or grass areas)~\cite{depth_and_flat}. Finally, other approaches propose to split the image into small tiles, and to classify them into different categories. In~\cite{lstm_tile_classif}, tiles are classified into water, trees, grass, bitumen or building categories, whereas in~\cite{manual_labeling_high_altitude} and \cite{kubernetes}, image patches are manually labeled and a CNN is trained to predict their suitability for landing.

The previous paragraph illustrates that current EL research lacks well-defined, unified objectives. Although the common underlying goal in most papers is to avoid hurting people and damaging infrastructures, a formal definition of these goals is missing. For example, while transportation infrastructures are considered dangerous for landing by some papers~\cite{parachute_from_database}, others consider that flat areas, such as roads, are safe landing spots~\cite{ELspot, depth_and_flat}. 
To address this issue, a recent paper conducted a hazard analysis for urban EL, and used it to define specific requirements for EL in the style of the SORA~\cite{ssiv}. This work concludes that a successful EL approach should avoid roads at all costs, and should not fall near pedestrians without a proper mitigation strategy to reduce the kinetic energy of the drone (e.g., parachute). In addition, The EL module should be tested extensively under different kinds of realistic potential threats (e.g., camera defect, meteorological conditions). Finally, runtime monitoring techniques should be implemented to ensure proper behavior of complex functions based on ML. These requirements are the guiding principles of both the design and test phases of the EL approach presented in this paper.

\subsection{Runtime Monitoring of ML} \label{sec:literature_rm}

Large ML models have introduced efficient solutions to computer vision problems that had long remained unsolved, including safety critical operations such as urban EL. 
However, due to their 
high complexity, conventional dependability techniques cannot guarantee safety~\cite{mlsafety}. To gain confidence in these systems, new approaches are 
being developed to monitor ML components during execution, so as to detect uncertain behavior and trigger appropriate safety actions.

A complete literature review about MLRM mechanisms is not the purpose of this paper, but we present a short taxonomy of existing techniques to illustrate the general philosophy inspired from~\cite{benchmark_raul}. A first family of approaches predicts misbehavior of a ML model by analyzing its inputs. For instance, reconstruction errors from auto encoders can be used to determine if an input is within the validity domain of the model~\cite{runtime_eval, runtime_autonomous_simu, RM_taxiing}. 
Another kind of Neural Network (NN) monitor models activation patterns of internal layers during training, and uses it to detect out-of-distribution (OOD) data~\cite{runtime_activation_patterns}. 
Finally, other approaches focus on the NN outputs to detect OOD data at runtime. For example, the highest softmax activation can be used as an anomaly score~\cite{softmax_OOD}, and when possible, softmax values can even be calibrated using OOD data~\cite{odin}. 
We also mention that some works have implemented MLRM mechanism within the context of practical safety applications~\cite{RM_taxiing, runtime_autonomous_simu, runtime_robotics_survey}. Similarly, our work studies the performance of three MLRM mechanisms within the context of a full EL architecture.

Regarding evaluation, several benchmark papers have already been published to assess the performance of different MLRM techniques.  For instance, the work presented in~\cite{runtime_eval} uses a benchmarking methodology based on three datasets, one for training the model, one for tuning the monitor and one for evaluation. Likewise, a complete benchmarking framework, relying on artificial generation of OOD data, is presented in~\cite{benchmark_raul}. These approaches present the advantage of using different data sources to fit a monitor and to assess its performance. However, they do not study the behavior of monitors at system level. This paper addresses this limitation by introducing an evaluation pipeline that can be applied to critical systems embedding MLRM, allowing to measure the safety benefits of using such mechanisms.




\section{EMERGENCY LANDING APPROACH} \label{sec:approach}

This work aims at finding safe landing zones in images collected from an on-board camera at high altitude ($\sim 50$m). According to~\cite{ssiv}, 
a valid approach for safe EL is to:
\begin{enumerate}
    \item Select an area far from busy roads.
    \item Go over this area and open a parachute.
\end{enumerate}
Such an implementation with good assurance would guarantee that no fatality will occur. In this work, buildings are also not considered safe. Indeed, crashing into a facade can lead to catastrophic loss of control, and our approach cannot distinguish them from roofs. We note that this paper does not deal with minimizing the distance to the landing area, although this might be an interesting extension. 

\begin{figure}
    \centering
    \includegraphics[width=0.48\textwidth]{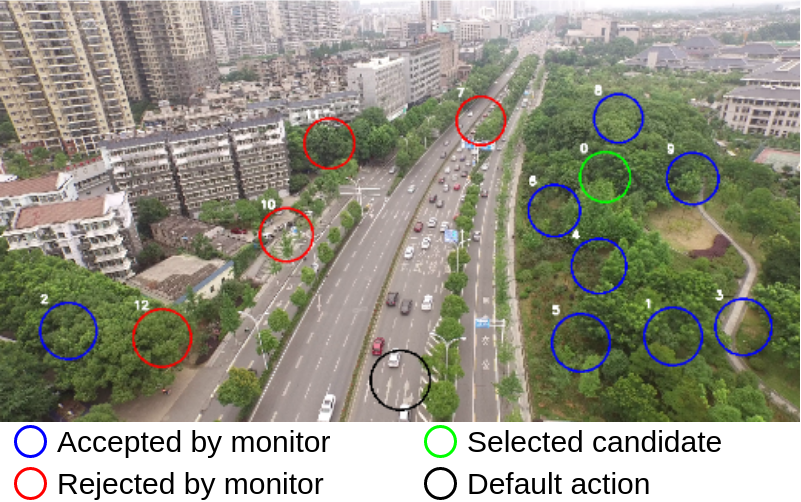}
    \caption{\textbf{Example of EL outputs}. Circles are candidates identified by the CM, colors indicate RM status. The default action consists in opening a parachute without moving.}
    \label{fig:example}
\end{figure}

The proposed architecture is composed of two main modules (Figure~\ref{fig:overview}). First, the \textit{Core Model} (CM) processes the image to identify candidate areas and ranks them based on a custom landing quality score. 
Next, the candidates are validated by the \textit{Runtime Monitor} (RM), supervising the ML components of the CM. 
In practice, 
if the best candidate is validated, 
the UAV goes to the selected area. Else, the candidate is discarded and the second best candidate goes through the monitoring step. This process continues until either a valid landing zone is found or all candidates are rejected. 
The outputs produced by the CM, and their associated monitoring status are illustrated in Figure~\ref{fig:example}. We underline that monitors are not fault free and can sometimes reject valid landing zones (see candidate 12 in Figure~\ref{fig:example}). The complete code for our approach is available~\cite{project_page}.

The scope of validity of our approach is defined by the specific settings of the UAVid dataset~\cite{uavid}, which was used for development and testing. UAVid contains 200 images for training and 70 for validation, extracted from 27 video sequences collected in different urban environments at a height of around 50 meters. The camera angle is set to 45 degrees (oblique view) and the images are high resolution~(4K). All 270 images are densely labelled for SemSeg, i.e., each pixel of each image is labelled with one of the following categories: building, road, static car, tree, low vegetation, humans, moving car, and background.

\subsection{Core Model}\label{sec:approach_core}

Once the EL system is triggered, an image $I$ is collected with the on-board camera. The CM processes $I$ and identifies $n$ candidate areas $\{c_1, ..., c_n\}$, which are ranked based on a custom landing quality score. This section presents the different steps composing the CM.

\subsubsection{Semantic segmentation}\label{sec:core_step1}

The image $I$ is first semantically segmented using a Multi-Scale Dilation network (MSDnet) trained on UAVid's training set. MSDnet is a large CNN model, which obtained the best results on UAVid~\cite{uavid}. Processing 4K images using MSDnet is prohibitively slow for such an emergency function ($\sim1$min). Hence, the original image is resized to 1024$\times$576, which reduces computation time significantly, while maintaining good segmentation results (visual evaluation). The HD image is used on small image patches as a MLRM mechanism (Section~\ref{sec:approach_runtime}).

\subsubsection{Valid candidates identification}\label{sec:core_step2}

A pixel of $I$ is considered valid for landing if no unsafe pixel (road, static or moving car, building) is present within a safety perimeter of radius $R$, dependent on UAV dimensions. To find the valid candidates, the segmentation map $I_{\text{seg}}$ is converted to a binary map representing the forbidden pixels $I_{\text{forbid}}$. Then, the ground dimension of a given pixel $I_{i,j}$ is estimated using optics formulas and geometric projections (detailed computation in~\cite{project_page}). 
To simplify computation, the ground dimension of each pixel belonging to row $i$ is considered equal, and noted $\delta_i$.
Then, the ground safety radius $R$ can be converted in pixels for any $I_{i, j}$: $\hat{R}_{i, j}=\beta.R/\delta_i$, where $\beta$ is a safety coefficient to account for the uncertainty on the UAV's height and camera angle. In our experiments, we use $R=2$~meters and set $\beta=1.7$, which guarantees that $R>2$~meters for error rates up to $20\%$~\cite{project_page}. 

In practice, to reduce processing time, $I_{\text{forbid}}$ is first divided into ten horizontal stripes of equal height. The two top stripes are removed as they contain the sky or unreachable areas (oblique view). 
For each remaining stripe, $\hat{R}_{i, j}$ is considered constant and is computed using the center row. This way, valid pixels are identified by simply convolving a square of side $\hat{R}_{i, j}$ on each stripe.

\subsubsection{Representative candidates selection}\label{sec:core_step3}

The number $N$ of candidates generated in the previous step is prohibitively high (up to tens of thousands). Thankfully, this huge list contains many neighboring pixels, representing similar areas. Hence, an algorithm is designed to retain only a small number $N^*$ (user defined) of representative candidates. 

To do so, the first step is to identify independent regions of connected candidate pixels. This is done by applying DBSCAN~\cite{dbscan} to the $N$ initial candidates, represented by their XY coordinates. DBSCAN is a powerful unsupervised algorithm to find clusters with high density of data points. This step generates $K$ clusters $\{\mathcal{C}_1, ..., \mathcal{C}_K\}$, and we note $N_k$ the number of candidates contained in $\mathcal{C}_k$.

Next, for each cluster $\mathcal{C}_k$, the objective is to select a number of candidate landing zones $N_k^*$ proportional to its size: $N_k^* = \max\left\{1, \text{round}\left(N^* \times N_k/N\right)\right\}$.
This way, each cluster has at least one representative, and $\sum_k N_k^* \approx N^*$. Next, the $N_k^*$ representatives of $\mathcal{C}_k$ are chosen so that their safe perimeters cover most of the cluster's area, while minimizing their mutual overlap. This is achieved by applying K-means clustering~\cite{kmeans} with $N_k^*$ centroids to the data points of $\mathcal{C}_k$. Fitting K-means to densely connected data points generates centroids that are spread out across the region, and can thus be used as representative candidates for the cluster. Finally, if there is too much overlap between the safe perimeters of two centroids, only one of them is retained. A max overlap of 25\% works well in practice to obtain a good coverage of the valid regions. 
The Scikit-learn implementations~\cite{scikit-learn} were used for both DBSCAN ($\epsilon=3$) and K-means (default parameters).

\subsubsection{Candidates ranking}\label{sec:core_step4}
The previous step generates the final set of $n=\sum_k N_k^*$ candidates $\{c_1, ..., c_n\}$. Then, the candidates are ranked in increasing order based on a custom hazard function $\mathcal{H}$. For a given candidate $c$, $\mathcal{H}$ is defined as:
\begin{equation}
    \mathcal{H}(c) = \alpha\times\mathcal{H}_s(c) + \left(1-\alpha\right)\times\mathcal{H}_d(c),
\end{equation}
where $\mathcal{H}_s(c)$ reflects the semantic categories present in $c$, $\mathcal{H}_d(c)$ the distance from $c$ to unsafe categories, and $\alpha \in [0, 1]$ is a weighting factor ($\alpha=0.5$ in our experiments, equal importance).

To define $\mathcal{H}_s$, a hazard score $s_l$ is associated with each authorized category (the lower, the safer). Then, $\mathcal{H}_s(c)$ is a weighted average of the pixels in the safe perimeter of $c$:
\begin{equation}\label{eq:s_sem}
    \mathcal{H}_s(c) = \frac{1}{\text{max}_l(s_l)} \times \sum_l m_l.s_l,
\end{equation}
where $m_l$ is the proportion of pixels in $c$ mapped to the $l^{\text{th}}$ category. $\mathcal{H}_s(c)$ ranges between $0$ and $1$. The values used in our experiments for the $s_l$'s are defined as follows:
Within the authorized categories, trees are considered more risky as the top view does not allow to know what hides below ($s_{\text{tree}}=3$). The background represents all objects from none of the 7 other categories. It is accepted because dangerous categories are supposed to be already removed. However, as we do not know precisely what it represents, it is given a higher score than the remaining categories ($s_{\text{background}}=2$). The presence of humans is not considered dangerous (parachute landing) ($s_{\text{human}}=1$) and low vegetation does not involve any hazard a priori ($s_{\text{vegetation}}=0$). The proposed scores are based on our understanding of the EL problem (see~\cite{ssiv} for more details regarding the safety analysis of EL), however, different implementation contexts might require different values based on specific hazard analyses.


On the other hand, $\mathcal{H}_d(c)$ is defined using $d(c)$, the distance between the center of $c$ and the closest forbidden pixel. 
If $d(c)$ exceeds a certain threshold $d_{\text{max}}$, then it is considered safe and $\mathcal{H}_d(c) = 0$ (we use $d_{\text{max}}=3R$). By construction, we know that $d(c) \geq R$, and thus $\mathcal{H}_d(c)$ can be brought to $[0,1]$ range as follows: 
\begin{equation}
    \mathcal{H}_d(c)=\begin{cases}
      (d(c)-d_{\text{max}})/(R-d_{\text{max}}), & \text{if}\ d(c) < d_{\text{max}} \\
      0, & \text{otherwise}.
    \end{cases}
\end{equation}


\subsection{Runtime safety monitoring strategies}\label{sec:approach_runtime}

Erroneous selection of a hazardous landing zone can have catastrophic consequences. Steps~\ref{sec:core_step2} to \ref{sec:core_step4} were tested empirically on the ground truth label maps of the UAVid training set, and no error was detected. This preliminary study suggests that the EL pipeline is reliable, as long as the segmentation maps produced by MSDnet are correct. 
Hence, to gain confidence in the safety of our EL approach, three strategies are proposed to monitor the predictions of MSDnet at runtime. They are chosen because they can be easily adapted to the SemSeg setting (most existing MLRM approaches are designed for classification).

\subsubsection{Local High Definition (LHD)}
To gain confidence in the system without excessively increasing processing time, we use the full resolution on local patches around the monitored candidates. This way, richer information is provided to MSDnet, and the improved segmentation helps to detect unsafe categories, and reject candidates accordingly. The other two monitors are built on top of LHD.

\subsubsection{Classification Hierarchy (CH)} 
\begin{figure}
    \centering
    \includegraphics[width=0.48\textwidth]{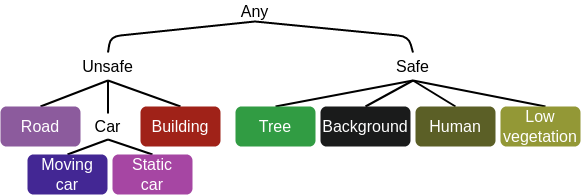}
    \caption{\textbf{Classification hierarchy}. The classification tree for emergency Landing corresponding to UAVid categories.}
    \label{fig:classif_hierarchy}
\end{figure}

The second MLRM mechanisms builds on the CFMEA safety analysis method~\cite{CFMEA}. First, a classification hierarchy is built to reflect the risk level of different MSDnet prediction uncertainties. For example, in the hierarchy presented in Figure~\ref{fig:classif_hierarchy}, if there is uncertainty between \textit{tree} and \textit{human}, the pixel is under-classified as \textit{safe}, i.e. the closest common ancestor. Similarly, uncertainty between \textit{road} and \textit{tree} results in under-classification as \textit{any}. This way, monitoring a candidate consists in under-classifying all of its pixels, and rejecting it if it contains either \textit{any} or \textit{unsafe} pixels. In our case, prediction uncertainty is assessed using a threshold $\tau$ on the softmax values, i.e., if several categories have softmax values greater than $\tau$, the result is uncertain among these categories. 
Several values of $\tau$ are tested in our experiments.

\subsubsection{Monte-Carlo Dropouts (MCD)}
The third monitoring technique leverages Monte-Carlo Dropouts~\cite{dropout_bayesian} (MCD) to improve uncertainty estimation. By maintaining dropout layers active at inference time (dropout rate $\rho$), MSDnet becomes non-deterministic, and running it on the same image multiple times enables to compute better statistical indicators of network uncertainty. In this work, the HD local patch around the monitored candidate is processed $N_{\text{MCD}}$ times by the stochastic MSDnet, and for each pixel, we note $\mu_i$ and $\sigma_i$ the empirical mean and standard deviation of the softmax value associated with category $i$. Then, the following scores are computed: $S_i^{\text{MCD}}=\mu_i + 3.\sigma_i$, 
and they are used instead of softmax to obtain pixel classification (or under-classification when relevant). If $\mathcal{N}(\mu_i, \sigma_i)$ is actually a good proxy for the distribution representing MSDnet confidence, there is over $99.7\%$ chances that the true value of the pixel's softmax score is below $S_i^{\text{MCD}}$. Several values of $\rho$ and $N_{\text{MCD}}$ are tested in our experiments.

\begin{figure*}
    \centering    
    \includegraphics[width=0.87\textwidth]{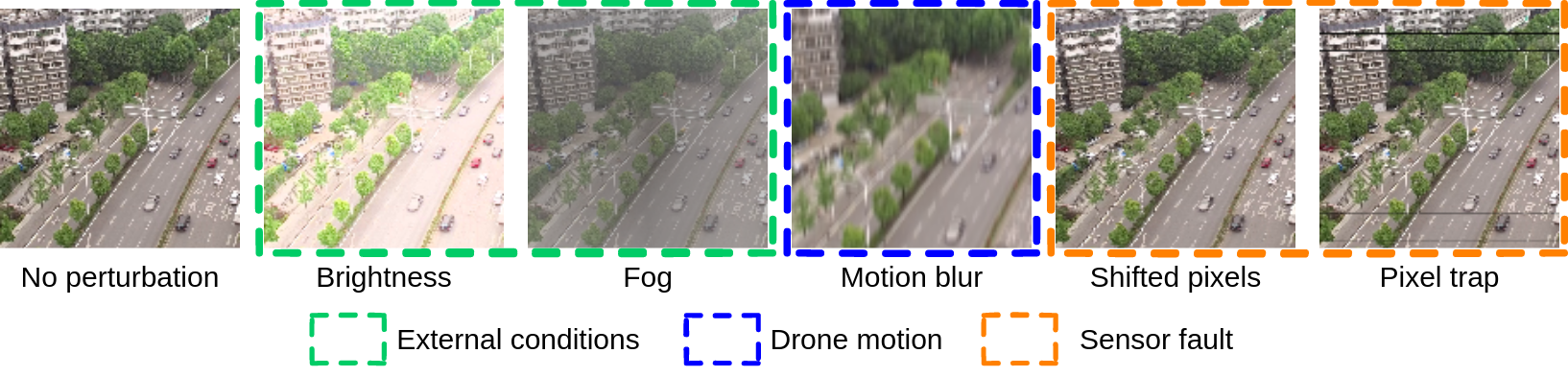}
    \caption{\textbf{Evaluation perturbations}. Examples to illustrate the synthetic image modifications used to evaluate the proposed Emergency Landing approach. The example image is from the UAVid validation set.}
    \label{fig:faults_examples}
\end{figure*}

\section{EVALUATION PROTOCOL} \label{sec:eval_rm}
This section presents a test protocol to evaluate the contributions of MLRM approaches to the safety of the whole system. It consists of three steps, inspired by FARM~\cite{farm}, a methodology to evaluate fault-tolerance mechanisms.

\subsection{Test scenarios}

The first step consists in defining relevant test scenarios with respect to potential failures of the applications. Namely, MLRM mechanisms must be evaluated on data representing live operations conditions, and accounting for realistic deviation from the nominal behavior. This step is highly context dependent, and can take various forms, such as generating simulation environments or collecting/labelling datasets.

To evaluate the proposed EL architecture, we use the validation set of UAVid. Its images present flight characteristics similar to the training set (drone height, camera angle, etc.), but are extracted from different video sequences, which allows to assess generalization of our EL approach to different locations. In addition, different kinds of realistic image perturbations are generated to gauge the benefits of using MLRM mechanisms against faulty inputs. Five perturbations are selected from~\cite{faults} in order to represent plausible sensor defects, changes in external conditions and failures related to UAV dynamic (Figure~\ref{fig:faults_examples}).   

\subsection{Readouts}

The system under test then runs on the evaluation scenarios defined above, and relevant execution variables – called readouts – are measured and stored in order to evaluate performance in the application context. To evaluate EL on a given image, 
the ground truth label map is used to compute the true hazard score $\hat{\mathcal{H}}(c)$ of each candidate $c$: 
\begin{equation}
\hat{\mathcal{H}}(c) = \begin{cases}
      1, & \text{if}\ c \supset \text{unsafe pixels},\\
      (1-\kappa)\times\hat{\mathcal{H}}_s(c), & \text{otherwise},
    \end{cases}
\end{equation}
where $\kappa$ defines a gap to separate candidates containing unsafe categories from others ($\kappa=0.5$ in our experiments), and $\hat{\mathcal{H}}_s(c)$ is computed using Eq.~\ref{eq:s_sem} with ground-truth labels. Using this formula forces candidates containing unsafe categories (even small areas) to have high hazard scores.

We also define the \textit{default action}, which consists in turning off the motors and opening a parachute without moving. This degraded safety action corresponds to what would be executed in case of failure, without the EL system. It is also the action executed if no candidate is accepted by the monitor. The bottom center area of the image is used as a proxy for the default landing area (Figure~\ref{fig:example}), as the exact spot is not present in the image (UAVid uses oblique view).

In summary, the following readouts are collected while running our test procedure on a given image:
\begin{enumerate*}
    \item the rank and true risk score of each candidate,
    \item the monitoring decision associated to each candidate,
    \item the true risk score of the default action,
    \item the monitoring times for each candidate.
\end{enumerate*}

\subsection{Metrics}
The readouts are then converted into metrics representing both the individual behavior of the MLRM mechanism under test, and the added safety resulting from its use.

\subsubsection{Monitor's individual behavior}
The role of a monitor is to assess the safety status of a candidate, i.e., safe or unsafe depending on whether it is free of unsafe categories. Hence, standard metrics from binary classification can be used to evaluate the performance of MLRM approaches. For EL monitoring, the following definitions hold: a \textit{true positive} is a rejected unsafe candidate, a \textit{false positive} is a rejected safe candidate, a \textit{true negative} is an accepted safe candidate, and a \textit{false negative} is accepted unsafe candidate.
Then, the following metrics are considered (range $[0,1]$):
\begin{itemize}
    \item \textbf{Precision} (P): 
    Proportion of rejected candidates that were actually unsafe. (High to avoid wasting 
    time rejecting safe candidates).
    \item \textbf{Recall} (R): 
    Proportion of unsafe candidates that were rejected. (High to avoid catastrophic events).
    \item \textbf{FP rate}: 
    Proportion of safe candidates that were rejected. (Low to avoid discarding all safe candidates).
\end{itemize}
Precision, recall and FP rate represent different dimensions under which the performance of a monitoring strategy can be evaluated. To simplify the interpretation of the results, two metrics aggregating performance scores are also reported:
\begin{itemize}
    \item \textbf{Matthews correlation coefficient} (MCC): 
    It accounts for all categories of the confusion matrix (range $[-1,1]$).
    \item \textbf{F1-score}: 
    Harmonic mean of P and R (range $[0,1]$).
\end{itemize}

\subsubsection{Monitor's added safety}
The safety added by a specific module of the EL pipeline is assessed by comparing the candidates chosen with and without this module. The difference between the two true hazard scores is called the \textit{safety gain}~$\mathcal{G}$. If we note $c_0$ the default candidate, $\Tilde{c}$ the candidate chosen by the CM and $c^*$ the one chosen by the full pipeline, we have:
\begin{equation}
    \mathcal{G}_{\text{CM}}=\hat{\mathcal{H}}(\Tilde{c}) - \hat{\mathcal{H}}(c_0),
\end{equation}
\begin{equation}
    \mathcal{G}_{\text{RM}}=\hat{\mathcal{H}}(c^*) - \hat{\mathcal{H}}(\Tilde{c}),
\end{equation}
\begin{equation}
    \mathcal{G}^*=\hat{\mathcal{H}}(c^*) - \hat{\mathcal{H}}(c_0) = \mathcal{G}_{\text{CM}} + \mathcal{G}_{\text{RM}}.
\end{equation}
These safety gains range between -1 and 1, and represent respectively the safety benefits brought by the CM, the additional benefits of using MLRM and the total benefits of the full architecture. To perform well under this metric, a MLRM mechanism should reject unsafe candidates (high recall), but should not be too conservative and reject too much safe candidates (low FP rate) or it might end up back to the default action and decrease CM results. 

Taking too much time to select the landing zone (low precision) can also be a source of hazard. Depending on the air traffic, it can increase the risk of collision or power failure. 
To evaluate monitoring time, the average overhead per image is also computed.

\begin{table*}
    \centering
    \caption{Results of the evaluation of the proposed EL system. Reported values are averaged across the entire evaluation dataset (70~images). Each perturbation is evaluated independently. Best results for each perturbation--metric pairs are in bold.}
    \label{tab:results}
    \resizebox{0.95\textwidth}{!}{
    \begin{tabular}{c|c|r|ccccccc|c}
    \Xhline{3\arrayrulewidth}
    \multirow{2}{*}{Perturbation} & \multirow{2}{*}{$\mathcal{G}_{\text{CM}}$} & \multicolumn{1}{c|}{\multirow{2}{*}{Monitor}} & \multirow{2}{*}{MCC} & \multirow{2}{*}{F$_1$} & \multirow{2}{*}{Precision} & \multirow{2}{*}{Recall} & \multirow{2}{*}{FP rate} & \multirow{2}{*}{$\mathcal{G}_{\text{RM}}$} & Overhead & \multirow{2}{*}{$\mathcal{G}^*$}\\ 
    & & & & & & & & & (sec) \\ \Xhline{3\arrayrulewidth}
    \multirow{4}{*}{None} & \multirow{4}{*}{0.502} & LHD & \textbf{0.711} & \textbf{0.814} & \textbf{0.813} & 0.816 & \textbf{0.105} & 0.144 & \textbf{2.12} & 0.646\\
    & & LHD+CH & 0.684 & 0.803 & 0.727 & 0.896 & 0.187 & 0.122 & 2.46 & 0.625 \\
    & & LHD+MCD & 0.708 & 0.813 & 0.810 & 0.816 & 0.107 & \textbf{0.148} & 34.53 & 0.650 \\
    & & LHD+CH+MCD & 0.676 & 0.797 & 0.708 & \textbf{0.912} & 0.209 & 0.132 & 36.40 & 0.634\\ \hline
    \multirow{4}{*}{Brightness} & \multirow{4}{*}{0.396} & LHD & \textbf{0.698} & 0.863 & \textbf{0.863} & 0.863 & \textbf{0.165} & 0.144 & \textbf{0.62} & 0.541 \\
    & & LHD+CH & 0.697 & \textbf{0.870} & 0.816 & 0.930 & 0.251 & \textbf{0.164} & 0.79 & 0.560 \\
    & & LHD+MCD & 0.695 & 0.862 & 0.861 & 0.863 & 0.168 & 0.144 & 37.13 & 0.541 \\
    & & LHD+CH+MCD & \textbf{0.698} & \textbf{0.870} & 0.807 & \textbf{0.944} & 0.271 & 0.143 & 33.75 & 0.539 \\ \hline
    \multirow{4}{*}{Fog} & \multirow{4}{*}{0.327} & LHD & 0.460 & 0.714 & \textbf{0.781} & 0.657 & \textbf{0.200} & \textbf{0.099} & \textbf{0.49} & 0.426 \\
    & & LHD+CH & 0.484 & 0.760 & 0.739 & 0.783 & 0.302 & 0.091 & 0.71 & 0.418 \\
    & & LHD+MCD & 0.456 & 0.712 & 0.779 & 0.657 & 0.203 & \textbf{0.099} & 28.94 & 0.426 \\
    & & LHD+CH+MCD & \textbf{0.485} & \textbf{0.767} & 0.726 & \textbf{0.813} & 0.334 & 0.067 & 30.07 & 0.394 \\ \hline
    \multirow{4}{*}{Motion blur} & \multirow{4}{*}{0.321} & LHD & 0.464 & \textbf{0.712} & \textbf{0.573} & 0.941 & \textbf{0.505} & \textbf{0.026} & \textbf{0.59} & 0.347 \\
    & & LHD+CH & \textbf{0.469} & \textbf{0.712} & 0.564 & \textbf{0.966} & 0.537 & 0.016 & 0.64 & 0.337 \\
    & & LHD+MCD & 0.464 & \textbf{0.712} & \textbf{0.573} & 0.941 & \textbf{0.505} & \textbf{0.026} & 33.35 & 0.347 \\
    & & LHD+CH+MCD & 0.466 & 0.711 & 0.563 & \textbf{0.966} & 0.541 & 0.008 & 23.83 & 0.329 \\ \hline
    \multirow{4}{*}{Shifted pixels} & \multirow{4}{*}{0.500} & LHD & \textbf{0.565} & 0.729 & \textbf{0.766} & 0.695 & \textbf{0.141} & \textbf{0.134} & \textbf{2.26} & 0.635 \\
    & & LHD+CH & 0.544 & 0.738 & 0.680 & 0.808 & 0.254 & 0.053 & 3.12 & 0.553 \\
    & & LHD+MCD & \textbf{0.565} & 0.730 & 0.764 & 0.699 & 0.143 & 0.124 & 36.37 & 0.625 \\
    & & LHD+CH+MCD & 0.543 & \textbf{0.739} & 0.672 & \textbf{0.821} & 0.267 & 0.053 & 42.23 & 0.553 \\ \hline
    \multirow{4}{*}{Pixel trap} & \multirow{4}{*}{0.519} & LHD & \textbf{0.671} & \textbf{0.797} & \textbf{0.744} & 0.858 & \textbf{0.170} & \textbf{0.087} & \textbf{2.57} & 0.607 \\
    & & LHD+CH & 0.616 & 0.765 & 0.664 & 0.903 & 0.263 & 0.055 & 3.31 & 0.574 \\
    & & LHD+MCD & \textbf{0.671} & \textbf{0.797} & \textbf{0.744} & 0.858 & \textbf{0.170} & \textbf{0.087} & 40.93 & 0.607 \\
    & & LHD+CH+MCD & 0.597 & 0.754 & 0.646 & \textbf{0.907} & 0.287 & 0.056 & 46.97 & 0.576 \\ \Xhline{3\arrayrulewidth}
    \end{tabular}}
\end{table*}

\section{RESULTS} \label{sec:results}

To the best of our knowledge, this work presents the first EL architecture addressing the requirements from~\cite{ssiv}. Other existing approaches were not designed to actuate in densely populated areas, and comparing them with our results using our evaluation criteria would be unfair. 

Regarding parameters of the different MLRM mechanisms, the following values were tested (chosen from preliminary experiments): $\tau \in \{0.25, 0.125\}$, $N_{MCD} \in \{10, 30\}$ and $\rho \in \{0.5, 0.75, 0.9\}$.
Results are presented in Table~\ref{tab:results}. To improve readability, only results for $\tau=0.25$, $N_{MCD}=10$ and $\rho=0.9$ are presented here, as they always perform better or similar to other values. F1-scores are not reported here as they behave similarly to MCC. The complete results can be found in~\cite{project_page}.

We first underline that all safety gains are positive, for all perturbations. This means that,  
on average, the core model chooses landing zones that are better than the default action, and that all MLRM mechanisms tested improve the final landing zone selected.

Comparing the performance of the proposed MLRM mechanisms, MCD appears to have little effect on the results, while involving a huge computational overhead. This might be due to the fact that MSDnet only contains two dropout layers, which breaks the fundamental assumption that each layer is followed by dropouts~\cite{dropout_bayesian}. Except for brightness, it also seems that adding CH actually decreases LHD performance. Finally, we note that good monitors usually have low FP rate (except brightness). On the other hand, monitors that are too conservative (better recall but worse FP rate) tend to have lower safety gain.

\section{CONCLUSIONS}

This paper presents a new approach to EL, which aims to address safety requirements identified in the recent literature. To this end, our architecture includes mechanisms to monitor ML components during execution. Hence, our work is also used as a practical use case to study the behavior of different MLRM approaches in the context of a real-world critical system. An extensive evaluation methodology is presented for generic systems embedding MLRM.

The results obtained suggest that existing MLRM mechanisms still lack robustness to ensure safety of critical ML systems. However, our work provides some insight about how to develop such technology. On the one hand, it shows that LHD alone contributes the most to improve EL safety, which suggests that successful monitors might be derived from richer information about the problem at hand. Even when some sensor information is too complex to cope with the main system implementation constraints, it might be useful for a monitor to focus on safety critical regions of the input space. On the other hand, it illustrates the importance of using system level metrics, such as the proposed safety gain, to properly assess the safety performance of MLRM mechanisms. Indeed, some monitors can perform well on individual metrics, but have limited interest for the whole system's safety (e.g., LHD+CH+MCD for shifted pixels).




\section*{ACKNOWLEDGMENT}

Our work has benefited from the AI Interdisciplinary Institute ANITI. ANITI is funded by the French "Investing for the Future – PIA3" program under the Grant agreement n°ANR-19-PI3A-0004.


\begin{thebibliography}{10}
\providecommand{\url}[1]{#1}
\csname url@rmstyle\endcsname
\providecommand{\newblock}{\relax}
\providecommand{\bibinfo}[2]{#2}
\providecommand\BIBentrySTDinterwordspacing{\spaceskip=0pt\relax}
\providecommand\BIBentryALTinterwordstretchfactor{4}
\providecommand\BIBentryALTinterwordspacing{\spaceskip=\fontdimen2\font plus
\BIBentryALTinterwordstretchfactor\fontdimen3\font minus
  \fontdimen4\font\relax}
\providecommand\BIBforeignlanguage[2]{{%
\expandafter\ifx\csname l@#1\endcsname\relax
\typeout{** WARNING: IEEEtran.bst: No hyphenation pattern has been}%
\typeout{** loaded for the language `#1'. Using the pattern for}%
\typeout{** the default language instead.}%
\else
\language=\csname l@#1\endcsname
\fi
#2}}

\bibitem{drones_urban_environment}
S.~Watkins, J.~Burry, A.~Mohamed, M.~Marino, S.~Prudden, A.~Fisher, N.~Kloet,
  T.~Jakobi, and R.~Clothier, ``Ten questions concerning the use of drones in
  urban environments,'' \emph{Building and Environment}, vol. 167, p. 106458,
  2020.

\bibitem{SORA}
{Joint Authorities for Rulemaking of Unmanned Systems ({JARUS})}, ``{JARUS}
  guidelines on {Specific Operations Risk Assessment} ({SORA}) v2.0,
  {Guidelines},'' 2019.

\bibitem{ssiv}
J.~Gu{\'e}rin, K.~Delmas, and J.~Guiochet, ``Certifying emergency landing for
  safe urban uav,'' in \emph{2021 50th Annual IEEE/IFIP International
  Conference on Dependable Systems and Networks Workshops (DSN-W), Safety and
  Security of Intelligent Vehicles (SSIV)}.\hskip 1em plus 0.5em minus
  0.4em\relax IEEE, 2021.

\bibitem{ELspot}
G.~Loureiro, A.~Dias, A.~Martins, and J.~Almeida, ``Emergency landing spot
  detection algorithm for unmanned aerial vehicles,'' \emph{Remote Sensing},
  vol.~13, no.~10, p. 1930, 2021.

\bibitem{low_altitude_flat}
L.~Chen, X.~Yuan, Y.~Xiao, Y.~Zhang, and J.~Zhu, ``Robust autonomous landing of
  {UAV} in non-cooperative environments based on dynamic time camera-{LiDAR}
  fusion,'' \emph{arXiv preprint arXiv:2011.13761}, 2020.

\bibitem{landmark_and_obstacle}
M.-F.~R. Lee, A.~Nugroho, T.-T. Le, S.~N. Bastida, \emph{et~al.}, ``Landing
  area recognition using deep learning for unammaned aerial vehicles,'' in
  \emph{2020 International Conference on Advanced Robotics and Intelligent
  Systems (ARIS)}.\hskip 1em plus 0.5em minus 0.4em\relax IEEE, 2020, pp. 1--6.

\bibitem{edges_svm}
L.~Mejias, ``Classifying natural aerial scenery for autonomous aircraft
  emergency landing,'' in \emph{2014 International Conference on Unmanned
  Aircraft Systems (ICUAS)}.\hskip 1em plus 0.5em minus 0.4em\relax IEEE, 2014,
  pp. 1236--1242.

\bibitem{edges_from_high_altitude_2}
L.~Mejias and D.~Fitzgerald, ``A multi-layered approach for site detection in
  {UAS} emergency landing scenarios using geometry-based image segmentation,''
  in \emph{2013 International Conference on Unmanned Aircraft Systems
  (ICUAS)}.\hskip 1em plus 0.5em minus 0.4em\relax IEEE, 2013, pp. 366--372.

\bibitem{edges_from_high_altitude_1}
L.~Mejias, D.~L. Fitzgerald, P.~C. Eng, and L.~Xi, ``Forced landing
  technologies for unmanned aerial vehicles: towards safer operations,''
  \emph{Aerial vehicles}, vol.~1, pp. 415--442, 2009.

\bibitem{depth_and_flat}
A.~Marcu, D.~Costea, V.~Licaret, M.~P{\^\i}rvu, E.~Slusanschi, and
  M.~Leordeanu, ``{SafeUAV}: Learning to estimate depth and safe landing areas
  for uavs from synthetic data,'' in \emph{Proceedings of the European
  Conference on Computer Vision (ECCV) Workshops}, 2018, pp. 0--0.

\bibitem{lstm_tile_classif}
K.~P. Lai, ``A deep learning model for automatic image texture classification:
  Application to vision-based automatic aircraft landing,'' Ph.D. dissertation,
  Queensland University of Technology, 2016.

\bibitem{manual_labeling_high_altitude}
I.~Funahashi, Y.~Umeki, T.~Yoshida, and M.~Iwahashi, ``Safety-level estimation
  of aerial images based on convolutional neural network for emergency landing
  of unmanned aerial vehicle,'' in \emph{2018 Asia-Pacific Signal and
  Information Processing Association Annual Summit and Conference (APSIPA
  ASC)}.\hskip 1em plus 0.5em minus 0.4em\relax IEEE, 2018, pp. 886--890.

\bibitem{kubernetes}
A.~Klos, M.~Rosenbaum, and W.~Schiffmann, ``Ensemble transfer learning for
  emergency landing field identification on moderate resource heterogeneous
  kubernetes cluster,'' \emph{arXiv preprint arXiv:2006.14887}, 2020.

\bibitem{parachute_from_database}
M.~Bleier, F.~Settele, M.~Krauss, A.~Knoll, and K.~Schilling, ``Risk assessment
  of flight paths for automatic emergency parachute deployment in {UAVs},''
  \emph{IFAC-PapersOnLine}, vol.~48, no.~9, pp. 180--185, 2015.

\bibitem{mlsafety}
G.~Schwalbe and M.~Schels, ``A survey on methods for the safety assurance of
  machine learning based systems,'' in \emph{10th European Congress on Embedded
  Real Time Software and Systems (ERTS 2020)}, 2020.

\bibitem{benchmark_raul}
R.~S. Ferreira, J.~Arlat, J.~Guiochet, and H.~Waeselynck, ``Benchmarking safety
  monitors for image classifiers with machine learning,'' in \emph{26th IEEE
  Pacific Rim International Symposium on Dependable Computing (PRDC 2021),
  Perth, Australia}.\hskip 1em plus 0.5em minus 0.4em\relax IEEE, 2021.

\bibitem{runtime_eval}
J.~Henriksson, C.~Berger, M.~Borg, L.~Tornberg, C.~Englund, S.~R.
  Sathyamoorthy, and S.~Ursing, ``Towards structured evaluation of deep neural
  network supervisors,'' in \emph{2019 IEEE International Conference On
  Artificial Intelligence Testing (AITest)}.\hskip 1em plus 0.5em minus
  0.4em\relax IEEE, 2019, pp. 27--34.

\bibitem{runtime_autonomous_simu}
A.~Stocco, M.~Weiss, M.~Calzana, and P.~Tonella, ``Misbehaviour prediction for
  autonomous driving systems,'' in \emph{Proceedings of the ACM/IEEE 42nd
  International Conference on Software Engineering}, 2020, pp. 359--371.

\bibitem{RM_taxiing}
D.~Cofer, I.~Amundson, R.~Sattigeri, A.~Passi, C.~Boggs, E.~Smith, L.~Gilham,
  T.~Byun, and S.~Rayadurgam, ``Run-time assurance for learning-based aircraft
  taxiing,'' in \emph{2020 AIAA/IEEE 39th Digital Avionics Systems Conference
  (DASC)}.\hskip 1em plus 0.5em minus 0.4em\relax IEEE, 2020, pp. 1--9.

\bibitem{runtime_activation_patterns}
C.-H. Cheng, G.~N{\"u}hrenberg, and H.~Yasuoka, ``Runtime monitoring neuron
  activation patterns,'' in \emph{2019 Design, Automation \& Test in Europe
  Conference \& Exhibition (DATE)}.\hskip 1em plus 0.5em minus 0.4em\relax
  IEEE, 2019, pp. 300--303.

\bibitem{softmax_OOD}
\BIBentryALTinterwordspacing
D.~Hendrycks and K.~Gimpel, ``A baseline for detecting misclassified and
  out-of-distribution examples in neural networks,'' in \emph{International
  Conference on Learning Representations}, 2017. [Online]. Available:
  \url{https://openreview.net/forum?id=Hkg4TI9xl}
\BIBentrySTDinterwordspacing

\bibitem{odin}
\BIBentryALTinterwordspacing
S.~Liang, Y.~Li, and R.~Srikant, ``Enhancing the reliability of
  out-of-distribution image detection in neural networks,'' in
  \emph{International Conference on Learning Representations}, 2018. [Online].
  Available: \url{https://openreview.net/forum?id=H1VGkIxRZ}
\BIBentrySTDinterwordspacing

\bibitem{runtime_robotics_survey}
Q.~M. Rahman, P.~Corke, and F.~Dayoub, ``Run-time monitoring of machine
  learning for robotic perception: A survey of emerging trends,'' \emph{IEEE
  Access}, vol.~9, pp. 20\,067--20\,075, 2021.

\bibitem{project_page}
J.~Guerin, ``{ANITI - UAV Emergency Landing},''
  \url{https://jorisguerin.github.io/ANITI_UavEmergencyLanding}, 2021.

\bibitem{uavid}
Y.~Lyu, G.~Vosselman, G.-S. Xia, A.~Yilmaz, and M.~Y. Yang, ``Uavid: A semantic
  segmentation dataset for uav imagery,'' \emph{ISPRS Journal of Photogrammetry
  and Remote Sensing}, vol. 165, pp. 108--119, 2020.

\bibitem{dbscan}
M.~Ester, H.-P. Kriegel, J.~Sander, X.~Xu, \emph{et~al.}, ``A density-based
  algorithm for discovering clusters in large spatial databases with noise.''
  in \emph{kdd}, vol.~96, no.~34, 1996, pp. 226--231.

\bibitem{kmeans}
J.~MacQueen \emph{et~al.}, ``Some methods for classification and analysis of
  multivariate observations,'' in \emph{Proceedings of the fifth Berkeley
  symposium on mathematical statistics and probability}, vol.~1, no.~14.\hskip
  1em plus 0.5em minus 0.4em\relax Oakland, CA, USA, 1967, pp. 281--297.

\bibitem{scikit-learn}
F.~Pedregosa, G.~Varoquaux, A.~Gramfort, V.~Michel, B.~Thirion, O.~Grisel,
  M.~Blondel, P.~Prettenhofer, R.~Weiss, V.~Dubourg, J.~Vanderplas, A.~Passos,
  D.~Cournapeau, M.~Brucher, M.~Perrot, and E.~Duchesnay, ``Scikit-learn:
  Machine learning in {P}ython,'' \emph{Journal of Machine Learning Research},
  vol.~12, pp. 2825--2830, 2011.

\bibitem{CFMEA}
R.~Salay, M.~Angus, and K.~Czarnecki, ``A safety analysis method for perceptual
  components in automated driving,'' in \emph{2019 IEEE 30th International
  Symposium on Software Reliability Engineering (ISSRE)}.\hskip 1em plus 0.5em
  minus 0.4em\relax IEEE, 2019, pp. 24--34.

\bibitem{dropout_bayesian}
Y.~Gal and Z.~Ghahramani, ``Dropout as a bayesian approximation: Representing
  model uncertainty in deep learning,'' in \emph{international conference on
  machine learning}.\hskip 1em plus 0.5em minus 0.4em\relax PMLR, 2016, pp.
  1050--1059.

\bibitem{farm}
J.~Arlat, A.~Costes, Y.~Crouzet, J.-C. Laprie, and D.~Powell, ``Fault injection
  and dependability evaluation of fault-tolerant systems,'' \emph{IEEE
  Transactions on computers}, vol.~42, no.~8, pp. 913--923, 1993.

\bibitem{faults}
\BIBentryALTinterwordspacing
D.~Hendrycks and T.~Dietterich, ``Benchmarking neural network robustness to
  common corruptions and perturbations,'' in \emph{International Conference on
  Learning Representations}, 2019. [Online]. Available:
  \url{https://openreview.net/forum?id=HJz6tiCqYm}
\BIBentrySTDinterwordspacing

\end{thebibliography}
\end{document}